\documentclass[conference]{IEEEtran}
\usepackage{microtype}
\usepackage{adjustbox}
\usepackage{subcaption}
\usepackage{amsmath}
\begin{document}
\title{Energy-Efficient Spiking Recurrent  Neural Network for Gesture Recognition on Embedded GPUs}

%\author{Marzieh Hassanshahi Varposhti}
%\email{marzieh.hassanshahi@donders.ru.nl}
%\orcid{0000-0001-7703-6835}
%\affiliation{%
%  \institution{Donders Institute} % for Brain, Cognition and Behaviour}
%  \city{Nijmegen}
% \country{The Netherlands}
%}

%\author{Mahyar Shahsavari}
%\email{mahyar.shahsavari@donders.ru.nl}
%\affiliation{%
 % \institution{Donders Institute} % for Brain, Cognition and Behaviour}
 % \city{Nijmegen}
 %\country{The Netherlands}
%}

%\author{Marcel van Gerven}
%\email{marcel.vangerven@donders.ru.nl}
%\affiliation{%
%  \institution{Donders Institute} % for Brain, Cognition and Behaviour}
%  \city{Nijmegen}
% \country{The Netherlands}
%}
\author{\IEEEauthorblockN{Marzieh Hassanshahi Varposhti }
\IEEEauthorblockA{Donders Institute for Brain, \\ Cognition and Behaviour\\Radboud University\\ 
Nijmegen, The Netherlands\\
marzieh.hassanshahi@donders.ru.nl
}
\and
\IEEEauthorblockN{Mahyar Shahsavari}
\IEEEauthorblockA{Donders Institute for Brain,\\ Cognition and Behaviour\\Radboud University\\
Nijmegen, The Netherlands\\
mahyar.shahsavari@donders.ru.nl}
\and
\IEEEauthorblockN{Marcel van Gerven}
\IEEEauthorblockA{Donders Institute for Brain,\\ Cognition and Behaviour\\Radboud University\\
Nijmegen, The Netherlands\\marcel.vangerven@donders.ru.nl}}

\maketitle
\begin{abstract}
 Implementing AI algorithms on event-based  embedded devices enables real-time processing of data, minimizes latency, and enhances power efficiency in edge computing. This research explores the deployment of a spiking recurrent neural network (SRNN) with liquid time constant neurons for gesture recognition. We focus on the energy efficiency and computational efficacy of NVIDIA Jetson Nano embedded GPU platforms.
 %specifically the NVIDIA Jetson Nano.
% Our comparative analysis extends to high-performance computing environments, utilizing an RTX 3000 Ada series GPU. By leveraging the advanced computing capabilities of PyTorch and CUDA for batch processing, we systematically assess throughput, frame rate, accuracy, and power consumption to establish a performance benchmark for real-time gesture recognition tasks.
The embedded GPU showcases a 14-fold increase in power efficiency relative to  a conventional GPU, making a compelling argument for its use in energy-constrained applications.
%However, the RTX 3000 Ada GPU maintains a distinct advantage in raw computational throughput, demonstrating a 6fold improvement in frame rate at equivalent batch sizes.
The study's empirical findings also highlight that batch processing significantly boosts frame rates across various batch sizes while maintaining accuracy levels well above the baseline. 
These insights validate the SRNN with liquid time constant neurons as a robust model for interpreting temporal-spatial data in gesture recognition, striking a critical balance between processing speed and power frugality. 
%The findings advocate for the advancement of neuromorphic computing, steering future research toward refining RSNNs for embedded systems, where efficiency is as crucial as performance.
 \end{abstract}

%\keywords{}
\IEEEpeerreviewmaketitle

\section{Introduction}
The advent of spiking neural networks (SNNs) implemented on edge devices and embedded systems has heralded a new era in the field of artificial intelligence and neuromorphic computing,  offering a more biologically plausible model of neural processing compared to traditional artificial neural networks (ANNs)~\cite{ARRAY2023}. Among various architectures, spiking recurrent neural networks (SRNNs) stand out for their capacity to process temporal sequences, making them particularly suited for dynamic tasks such as gesture recognition~\cite{yin2020effective}.

Gesture recognition is pivotal for human-computer interaction (HCI)~\cite{gesture_recognition_HCI}, necessitating both high accuracy and real-time processing. Traditional methods, relying on machine learning and deep learning, offer effective solutions, though at the cost of substantial computational resources. This poses a challenge for implementation on energy-constrained platforms like embedded systems.

The integration of dynamic vision sensor (DVS) ~\cite{DVS_refrence} cameras is a promising avenue to address these challenges. Unlike conventional cameras, DVS cameras only capture changes in brightness, significantly reducing data processing requirements. This feature makes them particularly suited for real-time, power-efficient gesture recognition tasks, enabling more effective deployment on devices where energy efficiency is crucial. By leveraging DVS cameras, gesture recognition systems can achieve faster processing times and reduced energy consumption, facilitating advanced HCI technologies on portable and wearable devices without compromising battery life~\cite{amir2017low}.

This paper elucidates the deployment of SRNNs employing liquid time constant (LTC) neurons~\cite{yin2021accurate} on embedded GPUs for gesture recognition, leveraging the PyTorch framework~\cite{paszke2019pytorch}.
The LTC neurons, known for their adaptive temporal response, significantly enhance the network's capability to process varied temporal dynamics, crucial for accurate gesture interpretation. Our approach leverages a pre-trained SRNN, trained with the forward propagation through time (FPTT) algorithm~\cite{yin2023accurate}. It is important to note that our focus is on the deployment aspect, particularly on embedded GPUs using PyTorch, rather than on novel neural network architectures or learning algorithms. That is, this work highlights the practical application and computational efficiency of SRNNs in real-time gesture recognition on resource-constrained embedded systems.

Our research is motivated by the need to develop energy-efficient, high-performance models for gesture recognition that can be deployed on embedded systems, such as the NVIDIA Jetson Nano. The use of embedded GPUs presents a unique set of challenges and opportunities in balancing computational speed with power consumption. By harnessing the inherent efficiencies of SRNNs and the dynamic capabilities of LTC neurons, our objective is to tackle these challenges, thereby expanding the frontiers of what can be achieved with embedded AI.

The contributions of this paper are twofold. First, we present a detailed analysis of the performance of an SRNN with LTC neurons for gesture recognition, trained using the FPTT algorithm. Second, we compare the energy efficiency and computational throughput of this network when deployed on an NVIDIA Jetson Nano versus an RTX 3000 Ada GPU. Our findings offer valuable insights into the trade-offs between power efficiency and processing speed, providing a benchmark for future research in neuromorphic computing and embedded AI.

\section{SNN on Embedded GPU}
In exploring the domain of real-time gesture recognition using SNNs on embedded platforms, particularly the NVIDIA Jetson Nano, it becomes evident that this area remains largely untapped, offering a fertile ground for pioneering research. The inherent energy efficiency and bio-inspired computation model of SNNs aligns well with the low-power, high-performance computing capabilities of the Jetson Nano, making it an ideal platform for deploying neuromorphic computing applications. However, challenges such as balancing model complexity with the computational constraints of embedded systems, achieving real-time processing speeds, and optimizing SNNs for deployment on non-native hardware frameworks such as CUDA must be addressed. This gap in the literature highlights a significant opportunity for innovation in real-time, energy-efficient gesture recognition systems. Developing solutions that tackle these challenges not only advances the field of neuromorphic computing but also opens new possibilities for real-time HCI in portable and wearable technology, thus setting the stage for future work at this promising intersection of neuromorphic engineering and embedded AI.

While the exploration of SRNNs and convolutional SNNs (ConvSNNs) on embedded platforms is still emerging, with specific implementations and comparisons in their early stages, the field of neuromorphic computing on such platforms has seen notable implementations using other spiking neuron models. 
For instance, the Izhikevich model, known for its ability to reproduce the rich dynamics of biological neurons with computational efficiency, has previously been implemented on the Jetson Nano~\cite{snngpu} and FPGA~\cite{SNN_poets} showcasing the potential of SNNs in edge computing devices. These implementations serve as valuable references for the neuromorphic computing community, demonstrating the feasibility of executing complex neural models on energy-efficient, low-power hardware. The success of models like the Izhikevich neuron on the Jetson Nano not only underscores the platform's capability to support neuromorphic computing but also highlights the broader potential for future research and development in deploying more complex SNN architectures, such as SRNNs and ConvSNNs, for real-world applications. This evolving landscape of SNN implementations on embedded systems opens up exciting avenues for leveraging the inherent advantages of SNNs—such as low power consumption, sparse and efficient processing of temporal and spatial data—in portable, real-time computing applications.

\section{Baseline network}

\begin{figure}[t]
    \centering
    % First part
    \begin{subfigure}[b]{0.2\textwidth}
        \includegraphics[width=\textwidth]{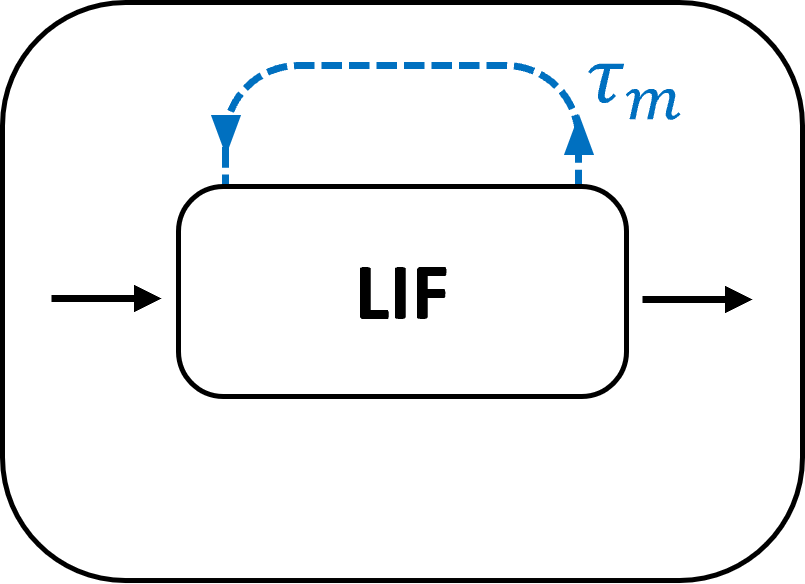}
        \caption{Leaky Integrate and Fire}
        \label{fig:partA}
    \end{subfigure}
    \hfill 
    % Second part
    \begin{subfigure}[b]{0.2\textwidth}
        \includegraphics[width=\textwidth]{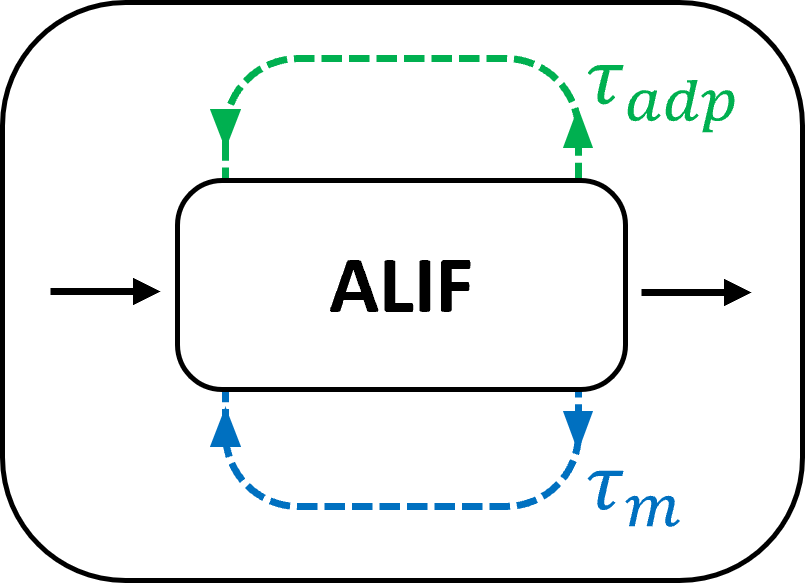}
        \caption{Adaptive LIF}
        \label{fig:partB}
    \end{subfigure}
    \hfill 
    % Third part
    \begin{subfigure}[b]{0.3\textwidth}
        \includegraphics[width=\textwidth]{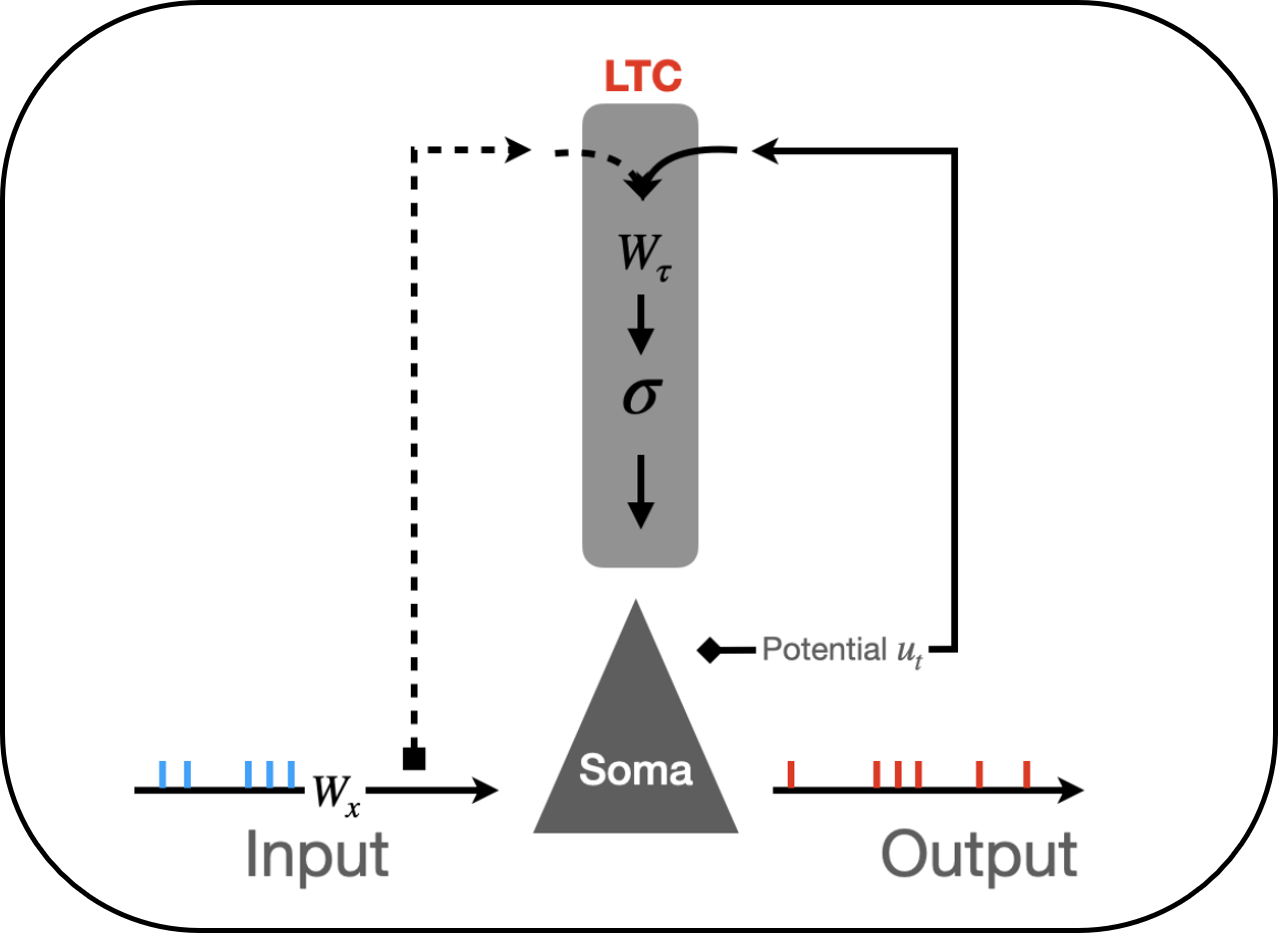}
        \caption{Liquid Time Constant }
        \label{fig:partC}
    \end{subfigure}
    
    \caption{Three common types of spiking neuron models.}
    \label{NEURONS}
\end{figure}

For our gesture recognition tasks, we implemented a baseline model which was trained using the FPTT algorithm~\cite{yin2023accurate}. %This approach is particularly fitting for our study, as it aligns with the real-time, dynamic nature of gesture recognition. 
The core of our baseline model is a SRNN featuring LTC neurons, which has shown exceptional proficiency in managing long temporal sequences and facilitating online learning, bypassing the limitations typically encountered when using backpropagation through time (BPTT) for training.

In SRNNs, network dynamics and learning performance are largely determined by the employed neuron model (see Fig.~\ref{NEURONS}). We start by describing the dynamics of conventional leaky integrate-and-fire (LIF) neurons, which are often employed to implement SNNs. The LIF neuron's membrane potential $u$ evolves in discrete time according to
%behavior is captured through a series of equations:
\begin{equation}
u_t = u_{t-1}\left(1 - \tau_m^{-1}\right) + \tau_m^{-1} R_m I_t \,.
\end{equation}
Here, the membrane potential $u$ of the neuron decays exponentially over time with a time constant $\tau_m$, influenced by the membrane resistance $R_m$ and the incoming signal 
\begin{equation}
    I_t = \sum_{t_i} w_i \delta(t_i) + I^\textrm{ext}_t \,,
\end{equation}
where $w_i$ represents the synaptic weight of the spike at times $t_i$, $\delta(t_i)$ denotes the Dirac delta function, signifying spike events at times $t_i$ and $I^\textrm{ext}_t$ represents externally injected current.

The neuron fires a spike $s_t$ when its membrane potential $u_t$ surpasses a threshold $\theta$. This process is modeled by the function
\begin{equation}
    s_t = f_s(u_t, \theta) \,, 
\end{equation}
which outputs 1 when a spike occurs and 0 otherwise. %The surrogate gradient $\hat{f}'_s(u_t, \vartheta)$ approximates this mechanism for computational efficiency. 
When a neuron exceeds the spike threshold, the membrane potential is reset according to 
\begin{equation}
u_t=u_t(1-s_t) + u_r s_t \,.
\label{eq:reset}
\end{equation}
The reset potential $u_r$ is considered to be zero in this model. 
%-----------------------------------------------------------------new section for writing the ALIF as it evolves from LIF------------------------

The performance of the LIF model can be enhanced by incorporating a more sophisticated, adaptive LIF (ALIF) neuron model~\cite{bellec2018long}. The model is determined by a coupled system of difference equations
%In these models, the threshold for firing spikes is raised after each spike and subsequently decays exponentially over time, with a decay constant. for implementing such a model, the parameters are as follows:
\begin{align*}
    u_t &= \alpha u_{t-1} + (1 - \alpha)R_m I_t - \theta_t s_{t-1}\\
    \eta_t &= \rho \eta_{t-1} + (1 - \rho)s_{t-1}
\end{align*}
where $\alpha = \exp(-\Delta t / \tau_m)$ and $\rho = \exp(-\Delta t / \tau_{adp})$
govern the temporal dynamics of the membrane potential and the threshold dynamics using time constants $\tau_m$ and $\tau_{adp}$. The threshold is dynamically adjusted using
\begin{equation}
    \theta_t = b_0 + \beta \eta_t
\end{equation}
with base level $b_0$ and an adaptive component proportional to $\eta_t$. The degree of threshold adaptation is determined by $\beta$. We use defaults of $b_0 = 0.1$ and $\beta = 1.8$  for the adaptive neurons.

\begin{figure*}[t]
    \centering
    \includegraphics[width=\textwidth]{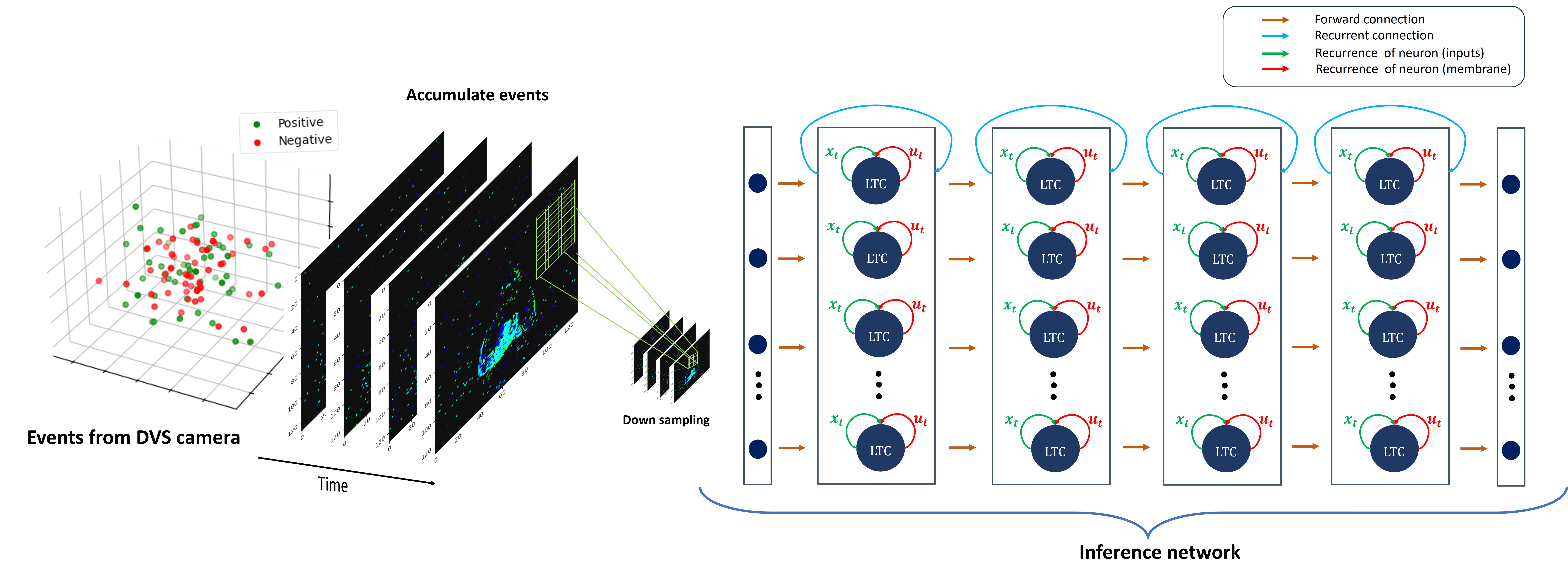}
    \caption{Inference procedure. a) preprocessing the raw data. b) Inference network including four spiking recurrent neural layers.}
    \label{infpro}
\end{figure*}
%---------------------------------------------------------------------end of the new section for ALIF------------------------
% Further extending this model, the adaptive leaky integrate-and-fire (ALIF) neuron includes an adaptation current $\omega$ into the neuron's state which, unlike the membrane potential $u$ is not reset after spiking, allowing it to retain longer memory compared to LIF neurons. The ALIF neuron is described by two coupled differential equations
% %incorporates threshold dynamics for adaptability:
% \begin{align}
% u_t &= \alpha u_{t-1} + (1 - \alpha)R_m I_t - \theta S_{t-1} \\
% \eta_t &= \rho \eta_{t-1} + (1 - \rho)S_{t-1} \\
% \theta &= b_0 + \beta \eta_t 
% %S_t &= \hat{f}_s(u_t, \vartheta).
% \end{align}
% Parameters $\alpha$ and $\rho$ govern the temporal dynamics of the membrane potential and the adaptive threshold, respectively, with $\alpha = \exp(-\Delta t / \tau_m)$ and $\rho = \exp(-\Delta t / \tau_{adp})$. The threshold $\vartheta$ dynamically adjusts, consisting of a base level $b_0$ and an adaptive component proportional to $\eta_t$, influenced by $\beta$, a constant determining the degree of threshold adaptation, set by default to 1.8 for adaptive neurons. 

%Int the SRNNs, the LIF spiking neuron is modeled as:
%\begin{align}
%u_{t-1} &= u_{t-1}(1 - S_{t-1}) + u_rS_{t-1} \\
%u_t &= u_{t-1}(1 - 1/\tau_m) + R_mI_t/\tau_m \\
%S_t &= f_s(u_t, \theta)
%\end{align}
%\section{The Liquid Spiking Neuron Model}
In this work, we employ the liquid time constant (LTC) neuron model~\cite{hasani2021liquid}, which generalizes the ALIF model and has been shown to enable the use of forward propagation through time (FPTT) in spiking neural networks (SNNs)~\cite{yin2023accurate}. The LTC neuron model equips SNNs with time-constants that are functions of the inputs and hidden states of the network. Also see~\cite{Quax2019} for an analysis of the functional benefits of using adaptive time-constants.

As illustrated in Fig.~\ref{fig:partC}, the LTC model extends the ALIF model by assuming that the membrane and adaptation time constants are determined by a sigmoid function $\sigma(z)$ of the input $z$. In case of the membrane time constant $\tau_m$, the input is given by the output of a (dense or convolutional) neural network layer consisting of weights $W_x$ and $W_\tau$ acting on $[x_t, u_{t-1}]$ with $x_t$ the neural network input at time $t$.  
Similarly, in case of the adaptation time constant $\tau_{adp}$, the input is given by the output of a (dense or convolutional) neural network layer acting on $[x_t, \eta_{t-1}]$ (more details can be found in  Appendix~\ref{appendix}). 

% \begin{equation}
% \tau^{-1} = \sigma(\mathbf{x}_t, \mathbf{u}_{t-1} \vert \mathbf{W}_\tau ), \label{eq:liquid_time_constant}
% \end{equation}
% where $\sigma$ is a sigmoid function used to ensure the inverse time-constant undergoes smooth transitions during learning. This liquid characteristic of time-constants, is fundamental to the liquid spiking neuron's behavior in this model.
% The liquid time-constants for standard adaptive spiking neurons are calculated either through a dense function for non-convolutional networks:
% \begin{equation}
% \tau^{-1}_m = \sigma(\text{Dense}([\mathbf{x}_t, \mathbf{u}_{t-1}]) ),
% \end{equation}
% or a convolution operation for spiking convolutional networks:
% \begin{equation}
% \tau^{-1}_m = \sigma(\text{Conv}(\mathbf{x}_t + \mathbf{u}_{t-1}) ).
% \end{equation}

The incorporation of LTC neurons into our SRNN model allows it to capture the complex spatio-temporal patterns essential for accurate gesture recognition from DVS data. By adapting the time constants in response to incoming stimuli, this SRNN can model the temporal dependencies with greater accuracy, which is particularly advantageous for real-time processing on energy-constrained edge devices.

In assessing the efficiency of the spiking neural network (SNN) methodology, an alternative approach involves transforming the SNN into a comparable recurrent neural network (RNN) by utilizing the ReLU activation function. This conversion is achieved by conveying the membrane potential to neighboring neurons at each time-step instead of the intermittent spike events, substituting equation~\eqref{eq:reset} with:
\begin{equation}
    s_t = \text{ReLU}(u_t - \theta_t).
\end{equation}
This adaptation allows for a direct comparison between the SNN and RNN models, facilitating a clearer understanding of the SNN's performance and applicability within our framework. %discussed in the baseline section of our paper.

To substantiate model's efficacy, we include visual representations of its performance metrics, such as accuracy and power consumption graphs, which effectively communicate the advantages of our baseline network over conventional methods. These visual aids will offer a clear, quantifiable comparison between the energy efficiency and computational throughput of the embedded and high-end GPU systems.

% The FPTT-trained LTC-SNN demonstrates a revolutionary advancement in computational efficiency and learning dynamics for SNNs. These networks are adept at handling the complexities of temporal information processing, showcasing a remarkable capability to learn from streaming input data efficiently. Such proficiency is evidenced by their performance on benchmark datasets, including the DVS Gesture dataset, where the FPTT-trained LTC-SNNs significantly outperformed their BPTT-trained counterparts over a range of sequence lengths, illustrating their superior adaptability and robustness.

% In our deployment, the baseline SRNN trained with FPTT exhibited a consistent accuracy profile, effectively capturing the spatio-temporal patterns essential for accurate gesture recognition. The network's architecture allowed for real-time learning and inference, tapping into the inherent advantages of SNNs for neuromorphic computing. Additionally, the network's training and operation on the NVIDIA Jetson Nano—an embedded GPU—proved to be energy-efficient.

% It's worth noting the baseline network's performance provides a benchmark against which we can measure the enhancements achieved through our findings. The FPTT framework's compatibility with the dynamics of LTC neurons suggests that our baseline model not only serves as a foundation for comparison but also as a potential springboard for future advancements in SNNs and gesture recognition systems.

\subsection{Advantage of LTC-based SRNNs}

Our choice to employ an LTC-based SRNN (LTC-SRNN)  is grounded in the need for a system that pairs the nuanced adaptability of biological neural processing with the pragmatic demands of embedded hardware efficiency. By adopting LTC-SRNNs, inspired by the liquid time constant concept~\cite{hasani2021liquid}, our model gains the ability to dynamically adjust its internal time constants in response to incoming signals, mirroring the functionality of LSTM gating but with enhanced temporal fidelity crucial for gesture recognition. % even after training for an adaptable network towards the input signals. 

This architectural choice directly addresses the prevalent challenge in deploying high-accuracy models within the power and processing constraints of embedded systems, such as NVIDIA's Jetson Nano. FPTT training of LTC-SRNNs emerges as a solution, significantly minimizing memory overhead while facilitating real-time, online learning capabilities. This allows our model to operate within the stringent energy budgets of embedded devices without sacrificing performance by having a shallow network while having the same performance. 
The demonstrated superiority of FPTT-trained LTC-SRNNs over traditional SNNs, particularly in handling long sequences across various benchmarks, underscores their potential, ensuring high model accuracy and responsiveness to temporal dynamics. % to redefine the standards of online learning and performance in SNNs. 
%The adaptability and efficiency of LTC-SNNs propel them as a front-runner in advancing gesture recognition technology, ensuring high model accuracy and responsiveness to temporal dynamics.

By integrating LTC-SRNNs trained via FPTT on embedded devices, our study not only addresses the limitations posed by embedded hardware but also %sets a new precedent for the development of complex, large-scale SNNs capable of 
also demonstrates how sophisticated tasks such as object detection can be achieved with unparalleled efficiency and low latency, 
%. This strategic integration signifies a leap towards the practical realization of neuromorphic computing's promise, 
offering a robust framework for enhancing real-time HCI applications with gesture recognition.

\section{Inference on AI edge device}

In this section, we explore the practical application of LTC-SRNNs for gesture recognition. The data used for this study originates from a DVS camera, which captures visual information in the form of events rather than conventional frames~\cite{amir2017low}. Unlike standard cameras that record redundant information at fixed intervals, the DVS camera provides an event-driven representation, capturing changes in intensity for each pixel asynchronously. This results in data that is inherently sparse, temporally precise, and offers high dynamic range, making it particularly suitable for capturing the nuanced movements involved in gestures (see Fig.~\ref{infpro}).

To utilize this event-based data for inference with our SRNN model, the events are accumulated over fixed-time intervals to create frames. This conversion process translates the asynchronous events into a format compatible with traditional temporal neural network architectures, while preserving the temporal resolution crucial for accurate gesture recognition. Different numbers of accumulated frames were tested during the model's inference phase to optimize for both speed and accuracy, allowing us to identify the optimal frame number that achieves the best balance between real-time processing requirements and the high accuracy necessary for reliable gesture interpretation.

The diagram provided in Fig.~\ref{workflow} succinctly captures the workflow of the experiment. Training the LTC-SRNN model is executed in PyTorch, a deep learning framework known for its flexibility and efficiency, which allows for rapid prototyping and supports CUDA for accelerated computation on GPUs~\cite{paszke2019pytorch}. Once trained, the model's parameters—including the weights, biases, and the dynamic time constants $\tau_m$ and $\tau_{adp}$ are preserved. This enables the trained LTC-SRNN to serve as a baseline for subsequent inference tasks.

The saved parameters form the foundation for the inference script, also implemented in PyTorch leveraging its CUDA integration. This script is designed to run the model efficiently on GPU architectures, exploiting the parallel processing capabilities GPUs provide. We then execute this script on two distinct hardware platforms: a local GPU, represented by a high-end workstation, and an embedded GPU, exemplified by the Jetson Nano which the software and hardware utilization are described in Table~\ref{featuretable}. The workstation GPU is expected to deliver higher frame rates due to its superior processing power, while the Jetson Nano is optimized for energy efficiency, crucial for edge computing applications.

\subsection{Inference process }

\begin{figure}[h]
    \centering
\includegraphics[width=0.85\columnwidth]{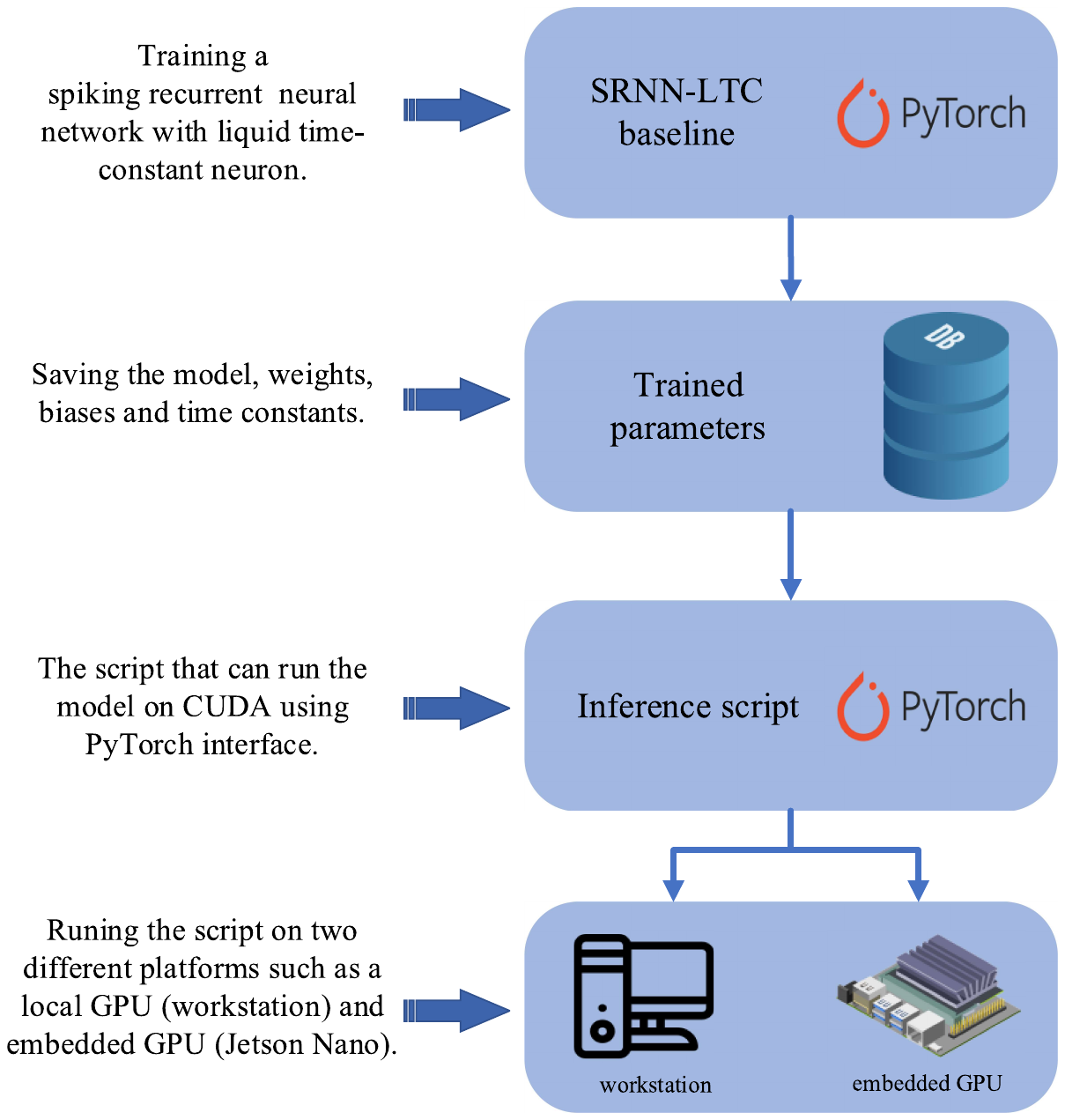}
    \caption{Experimental workflow.}
    \label{workflow}
\end{figure}

To realize inference on AI edge devices, our approach is structured into three stages:
\begin{enumerate}
\item 
\textbf{Initialization of trained network parameters:}
The first step involves loading the trained parameters that define the architecture and operational specifics of our SRNN model. This stage sets the foundation by reinstating the network's learned dynamics, ensuring it is primed for efficient inference.
\item 
\textbf{Baseline model integration:}
Subsequently, we incorporate the baseline model, as detailed in the preceding sections of this paper. This integration ensures that the inference process leverages the foundational SRNN model equipped with LTC neurons, characterized by its exceptional ability to handle temporal sequences with high precision and energy efficiency.
\item 
\textbf{Execution of inference script:}
 The final step entails running the inference script, as illustrated in Fig~\ref{workflow}. This script is meticulously designed to process input data through a series of computational steps, tailored to the unique requirements of our SRNN model.
\end{enumerate}
The inference script operates through a three-step process, detailed as follows:
\begin{itemize}
    \item 
    {\bf Event frame accumulation}: Given the SRNN model's inability to process raw event data directly, a crucial preprocessing step is employed. Here, raw events captured by the DVS camera are accumulated into T number of frames within fixed time intervals. This accumulation converts the asynchronous event data into a synchronous frame-based representation, preserving the temporal accuracy essential for gesture recognition \cite{preprocessingdata}.
    \item 
    {\bf Downsampling for optimization}: Processing frames at full (128 $\times$ 128 pixels) resolution is computationally demanding and may not be optimal for real-time inference on edge devices. To address this, frames are downscaled to a $32 \times 32$ pixel resolution. This downsampling significantly reduces the input size, making the inference process more efficient without substantially compromising the quality of gesture recognition.
    \item 
    {\bf Gesture classification via the baseline network}: The final step involves feeding the downsampled frames into the baseline SRNN model. The network then classifies the input gesture based on the spatiotemporal patterns encoded in the frames. This classification marks the culmination of the inference process, yielding the identified gesture as the output.
\end{itemize}

Our model training was fine-tuned to ensure peak accuracy by selecting the best-performing frame number. However, during inference, the flexibility of our approach allows for experimentation with various frame numbers to adapt to the specific requirements of the application and the constraints of the deployment environment. This dual-phase approach-- optimal training followed by adaptable inference -- ensures that the model can be finely adjusted to maintain a high standard of accuracy, even when real-time constraints necessitate a trade-off with frame rate.

In conclusion, our approach encapsulates the full cycle of developing and deploying a sophisticated neural network model for gesture recognition, from data capture using cutting-edge sensor technology to model training and adaptation for real-time inference on edge devices. The LTC-SRNN model's compatibility with both powerful workstations and energy-conscious embedded systems demonstrates its versatility and readiness for real-world applications, making it a prime example of the advances and adaptability in the domain of neuromorphic computing.
% \begin{figure}
%     \centering
%     \includegraphics[width=\columnwidth]{table1.png}
%         \caption{TABLE I: Features of the selected hardware for inference.}
%     \label{fig:featuretable}
% \end{figure}

\renewcommand{\arraystretch}{1.3} 
\begin{table}
\caption{Features of the selected hardware for inference.}
\label{featuretable}
\large
\resizebox{\columnwidth}{!}{%
\begin{tabular}{|p{0.2\linewidth}|p{0.4\linewidth}|p{0.4\linewidth}|}
\hline
\multicolumn{1}{|c|}{\textbf{Feature}}     & \multicolumn{1}{c|}{\textbf{Workstation}}                                                       & \textbf{Embedded GPU}                       \\ \hline
\multicolumn{1}{|c|}{Reference system}     & \multicolumn{1}{c|}{\begin{tabular}[c]{@{}c@{}}Dell precision 5480 \\ workstation\end{tabular}} & {\begin{tabular}[c]{@{}c@{}}Jetson Nano\end{tabular}}          \\ \hline
\multicolumn{1}{|c|}{CPU}                  & \multicolumn{1}{c|}{\begin{tabular}[c]{@{}c@{}}13th Gen  \\ Intel(R) Core(TM) i7-13800H\\ 2.5 GHz\end{tabular}} & {\begin{tabular}[c]{@{}c@{}}Quad-Core Arm®  \\Cortex®-A57\\ MPCore  1.43 GHz\end{tabular}}   \\ \hline
\multicolumn{1}{|c|}{Operating system}     & \multicolumn{1}{c|}{Windows 11}                                                                 & Ubuntu 20.04                                \\ \hline
\multicolumn{1}{|c|}{PyTorch version}      & \multicolumn{1}{c|}{2.2.0}                                                                      & 1.13.0                                      \\ \hline
\multicolumn{1}{|c|}{CUDA version}         & \multicolumn{1}{c|}{12.1}                                                                       & 10.2                                        \\ \hline                  
\multicolumn{1}{|c|}{GPU Device name}          & \multicolumn{1}{c|}{\begin{tabular}[c]{@{}c@{}}NVIDIA  \\ RTX 3000 Ada\\ Generation Laptop GPU\end{tabular}}& {\begin{tabular}[c]{@{}c@{}}NVIDIA Maxwell™ \\ Architecture\end{tabular}}\\ \hline
\multicolumn{1}{|c|}{Number of CUDA cores} & \multicolumn{1}{c|}{4,608}                                                                      & 128                                         \\ \hline
\multicolumn{1}{|c|}{GPU max frequency}    & \multicolumn{1}{c|}{1695 MHz}                                                                   & 921 MHz                                     \\ \hline
\multicolumn{1}{|c|}{GPU Memory}               & \multicolumn{1}{c|}{8 GB 128-bit GDDR6 16GB/s}                                                  & 4GB 64-bit LPDDR4 25.6GB/s                  \\ \hline
\end{tabular}%
}
\end{table}

\section {Experimental results}
The evaluation of spiking neural network (SNN) performance across various batch sizes and number of frames for every sequence ($T$) was conducted on two distinct hardware platforms: a standard GPU and an NVIDIA Jetson Nano embedded GPU. The primary metrics for assessment included frame rate (inferences per second) and normalized accuracy regarding the baseline model (percentage of correct inferences).

To ensure optimal performance measurement of the GPUs, a warm-up procedure was employed prior to the execution of batch processing experiments. This step is crucial as GPUs may not operate at their maximum clock speeds when not previously engaged in heavy tasks. By warming up the GPU, we guaranteed that the device was running at its full capability during our performance assessments, ensuring the accuracy of our comparative analysis.

\subsection{Inference on standard GPU}

The frame rate results for the standard GPU configuration in Fig.~\ref{Local GPU} demonstrate a clear correlation between batch size and processing speed, with larger batch sizes yielding significantly higher frame rates. For $T=20$, the frame rate steadily increases from 380 frames/sec for a batch size of 16 to 5150 frames/sec for a batch size of 1024. The trend is consistent across other $T$ values, while keeping the performance (accuracy) around the baseline model. 

It is notable that for higher $T$ values ($T=35$ and $T=50$), the accuracy metric will remain around the baseline, suggesting that the network's temporal processing capabilities do not hinder its inferential accuracy. A marginal decrease in accuracy is observed at $T=20$ and $T=100$ indicating a potential trade-off.

\subsection{Inference on Jetson Nano}

%\subsubsection{Analysis of batch processing performance}

\begin{figure*}[ht]
    \centering
    \begin{subfigure}{0.48\textwidth}
        \centering
        \includegraphics[width=\linewidth]{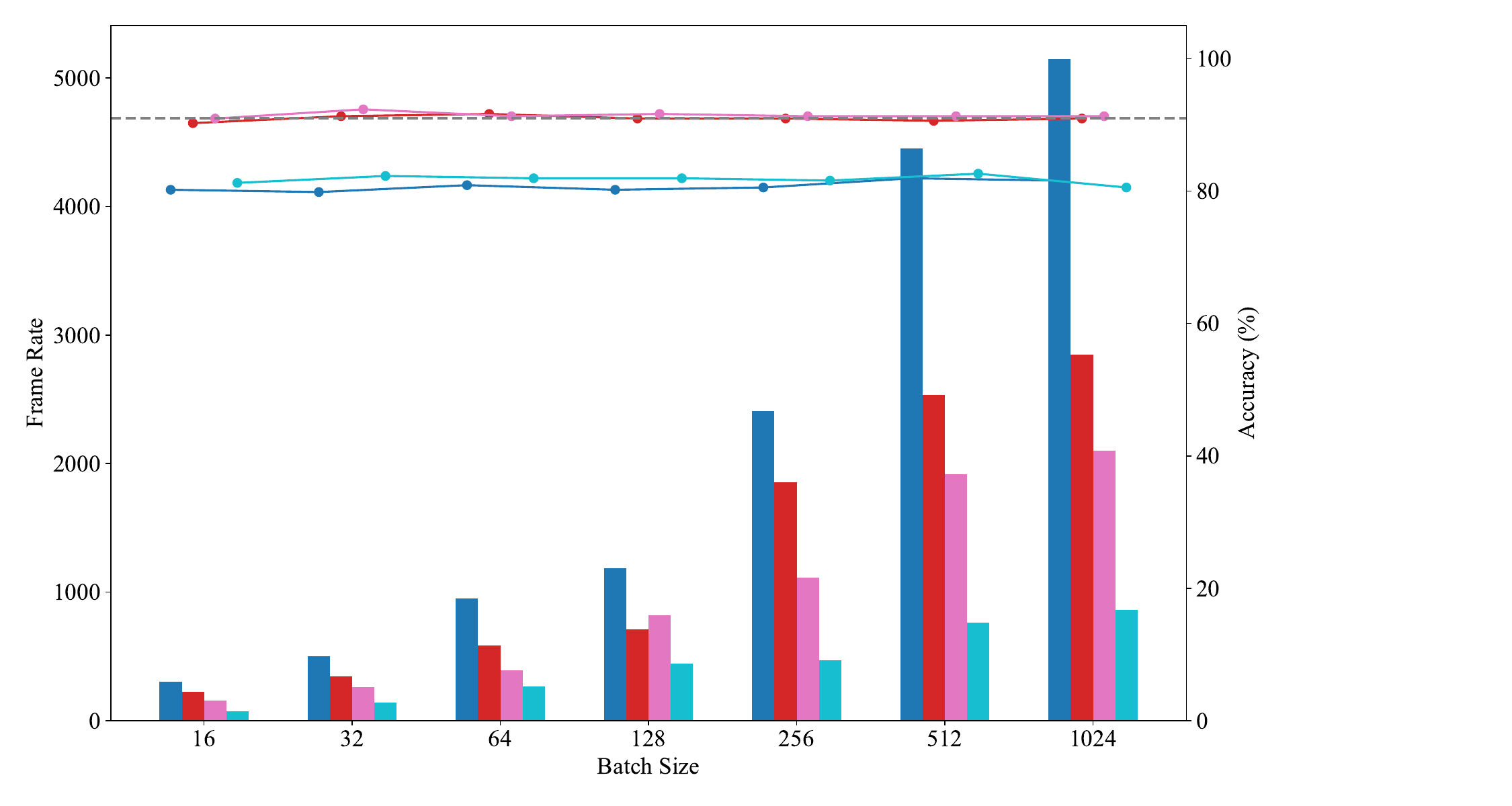}
        \caption{Local GPU}
        \label{Local GPU}
    \end{subfigure}\hfill
    \begin{subfigure}{0.48\textwidth}
        \centering
        \includegraphics[width=\linewidth,height=0.53\textwidth]{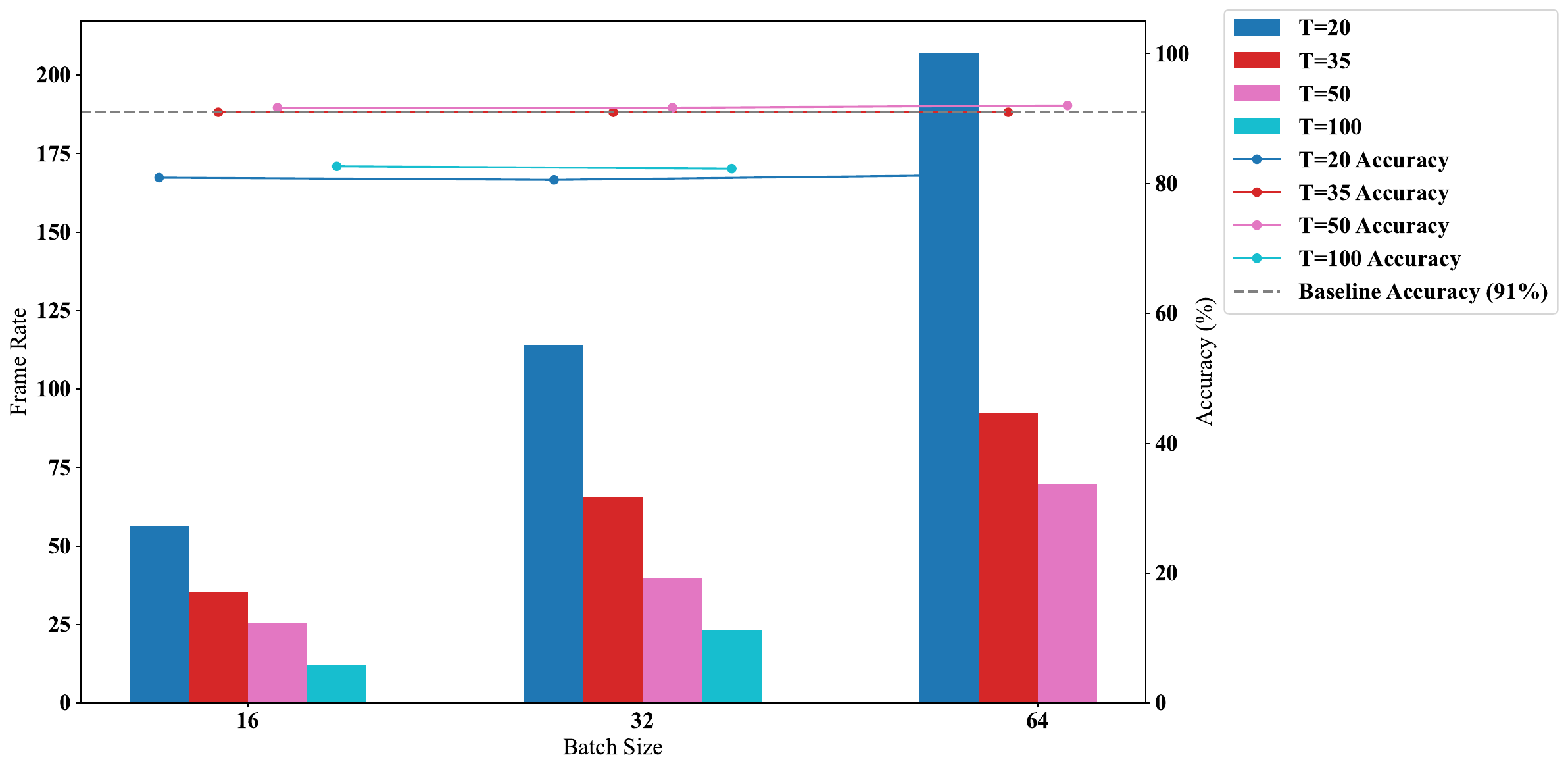} 
        \caption{Jetson Nano}
        \label{Jetson Nano}
    \end{subfigure}
    \caption{Batch processing experimental results. Panels compare frame rate and accuracy for different batch sizes and $T$ values when using (a) the local GPU or (b) the embedded GPU.}
    \label{fig:combined}
\end{figure*}

Figure~\ref{Jetson Nano} shows the results for the same experiment on the edge device (Jetson Nano). The data reveals that as the batch size increases, there is a significant increase in the frame rate for smaller $T$ values (20 and 35), whereas the frame rates for $T$ values of 50 and 100 remain relatively stable across all batch sizes. The largest batch size of 64, in particular, shows a remarkable spike in the frame rate for $T=20$ and $T=35$. This suggests that the Jetson Nano can process smaller accumulations of frames more efficiently, even with larger batch sizes.

Accuracy trends indicate that higher $T$ values yield more consistent accuracy results, albeit at a cost to frame rate. For $T=20$ and $T=35$, there is a visible drop in accuracy with increasing batch sizes, which is not as pronounced with $T=50$ and $T=100$. The baseline accuracy is maintained at $91\%$, and none of the configurations exceed this threshold. This pattern suggests that while smaller $T$ values can benefit frame rate, yet they sacrifice the accuracy.

These findings underscore the importance of configuring the frame accumulation parameter appropriately to balance the real-time processing requirements with the accuracy demands of applications running on edge devices such as the Jetson Nano.

\begin{figure}
    \centering
    \includegraphics[width=\columnwidth]{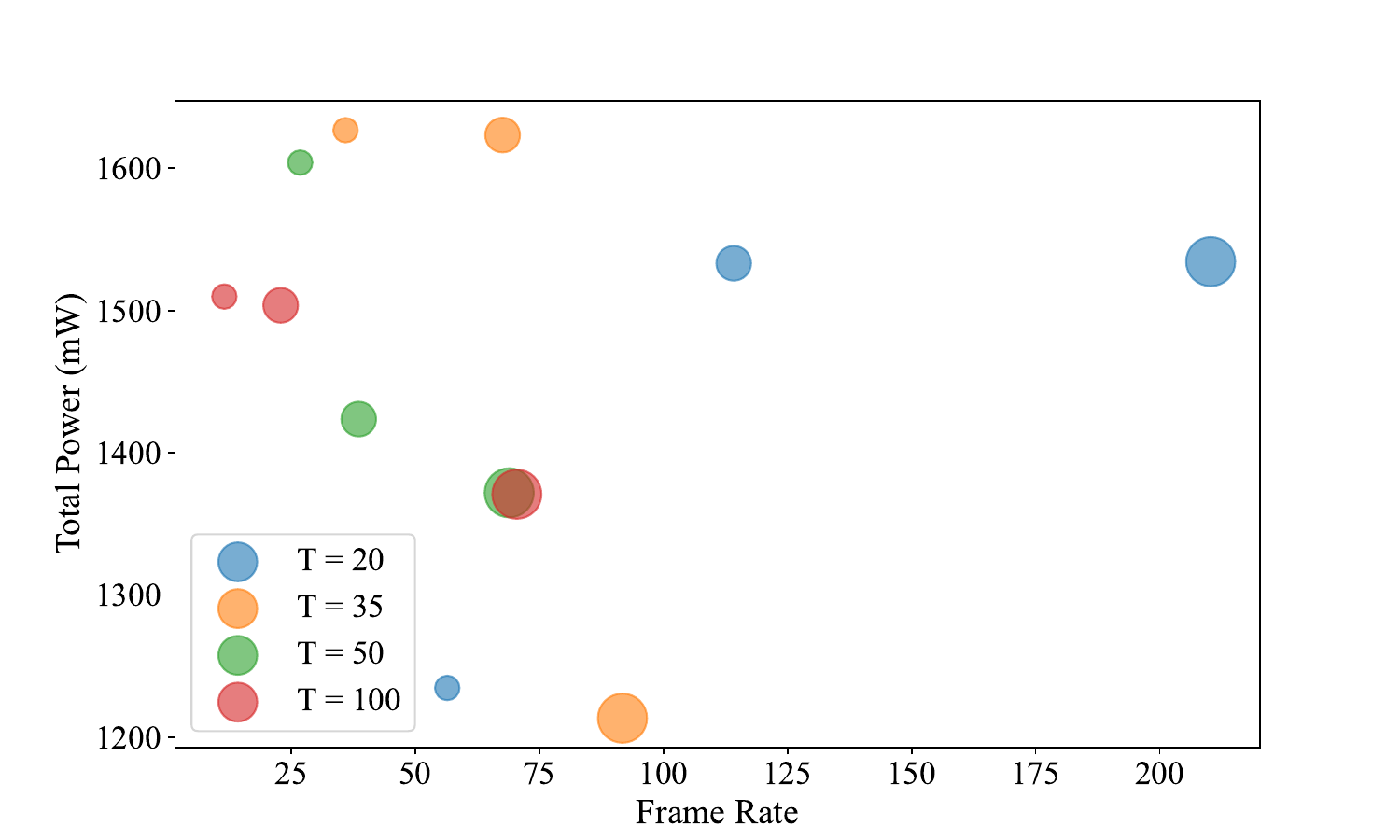}
    \caption{Power consumption of the Jetson Nano. Batch size indicated by node width.}
    \label{jetpower}
\end{figure}

% \subsubsection{GPU warm-up procedure}
% To ensure optimal performance measurement of the GPUs, a warm-up procedure was employed prior to the execution of batch processing experiments. This step is crucial as GPUs may not operate at their maximum clock speeds when not previously engaged in heavy tasks. By warming up the GPU, we guaranteed that the device was running at its full capability during our performance assessments, ensuring the accuracy of our comparative analysis.
Figures~\ref{Local GPU} and \ref{Jetson Nano} depict the results of our batch processing experiments, illustrating the relationship between batch size and both frame rate and accuracy. These figures highlight the GPU's ability to efficiently process larger batches, capitalizing on CUDA's parallel processing capabilities. Furthermore, the results demonstrate that despite the inherent constraints of the  Jetson Nano, our SNN model achieves commendable performance, balancing computational speed with energy consumption, a testament to the viability of deploying advanced neural network models on edge devices.

%\subsubsection{Power consumption measurement on Jetson Nano}
In aiming for precise quantification of power consumption, all non-essential tasks, including display outputs for monitors, were disabled on the Jetson Nano. Power consumption during the inference process was measured using a specific method involving the execution of the \textit{tegrastats} command through a Python subprocess. This method allowed for real-time monitoring of the device's power usage, providing valuable insights into the energy efficiency of our SNN implementation under varying computational loads.
This detailed approach to measuring power consumption reinforces the importance of energy efficiency in edge computing scenarios, particularly when deploying advanced neural network models like SNNs.  Power consumption results for the Jetson Nano are shown in Fig.~\ref{jetpower} considering different batch sizes in the format of the size of each node and $T$ numbers showing the comprehensive trade-off between the frame rate and the batch size for target power consumption. Table~\ref{compairpower} shows that the Jetson Nano can achieve its optimal performance at an order of magnitude less power compared to the local GPU setup. 
% Note that the frame/s is showing the capability of real-time processing, not the prediction. In general the number of frames processed by  the model affects the ability of real-time processing and performance of the application, as well as the provided power for edge computing. 

%\subsubsection{Analysis of batch processing performance}

\renewcommand{\arraystretch}{1}
\begin{table}[]
\caption{Comparison of the best performance for  each hardware platform with respect to the power consumed.}
\label{compairpower}
\resizebox{\columnwidth}{!}{%
\begin{tabular}{|c|c|c|}
\hline
\textbf{Device Type} & \textbf{\begin{tabular}[c]{@{}c@{}}Maximum frame/s\\ (accuracy)\end{tabular}} & \multicolumn{1}{l|}{\textbf{Average power consumption (W)}} \\ \hline
\begin{tabular}[c]{@{}c@{}}Embedded GPU\\ (Jetson Nano)\end{tabular} & \begin{tabular}[c]{@{}c@{}}155\\ 80\%\end{tabular} & 1.5 \\ \hline
\begin{tabular}[c]{@{}c@{}}Local GPU \\ (RTX 3000 Ada)\end{tabular} & \begin{tabular}[c]{@{}c@{}}5127\\ 80\%\end{tabular} & 21 \\ \hline
\end{tabular}%
}
\end{table}
\section{Conclusion}
Our investigation into deploying energy-efficient recurrent spiking neural networks with liquid time constant neurons for gesture recognition on edge devices, particularly the NVIDIA Jetson Nano, has shown promise for edge-based AI, reducing the energy need of current AI systems. We demonstrated the feasibility of conducting inference with high accuracy and efficiency on such devices.  The adoption of the FPTT algorithm to effectively train SRNNs played a pivotal role here. 
%, significantly reducing memory overhead compared to traditional methods, thus enabling continual learning directly on the edge. 
This capability opens up new possibilities for the deployment of adaptive spiking neural networks in real-time applications, marking a significant advancement in edge computing. 
%The spiking version of the networks showcases that with holding the performance of the model, it will be feasible to run them on more constrained hardware or eliminate the carbon footprint for more sophisticated and recursive applications such as Large Language Models (LLM) as shown a promising window in \cite{matmulfree}.
%\section{future work}

\section{ACKNOWLEDGEMENTS}

This publication is part of the project ROBUST: Trustworthy AI-based Systems for Sustainable Growth with project number KICH3.L TP.20.006, which is (partly) financed by the Dutch Research Council (NWO), ASMPT, and the Dutch Ministry of Economic Affairs and Climate Policy (EZK) under the program LTP KIC 2020-2023. All content represents the opinion of the authors, which is not necessarily shared or endorsed by their respective employers and/or sponsors.
We thank Bojian Yin for his useful comments and kindly making the FPTT code available to support this work. 

\bibliographystyle{IEEEtran}

%\bibliography{reference}
\appendix
\label{appendix}
\section{Appendix A}
The following equations delineate the update dynamics of a liquid spiking neuron:
% \begin{align*}
% \text{$\tau_{\text{adp}}$ update : } \rho &= \tau_{\text{adp}}^{-1} = \sigma([x_t, b_{t-1}] \mathbin\Vert W_{\tau_{\text{adp}}}) \\
% \text{$\tau_m$ update : } \tau_m^{-1} &= \sigma([x_t, u_{t-1}] \mathbin\Vert W_{\tau_m}) \\
% \text{$\theta_t$ update : } b_t &= \rho b_{t-1} + (1 - \rho) s_{t-1} \\
% \theta_t &= 0.1 + 1.8 b_t \\
% \text{$u_t$ update : } \Delta u &= (-u_{t-1} + x_t) / \tau_m \\
% u_t &= u_{t-1} + \Delta u \\
% \text{spike $s_t$ : } s_t &= f_s(u_t, \theta) \\
% \text{resting : } u_t &= u_t (1 - s_t) + u_{\text{rest}} s_t,
% \end{align*}
\begin{align*}
\rho &= \tau_{\text{adp}}^{-1} = \sigma([x_t, b_{t-1}] \mathbin\Vert W_{\tau_{\text{adp}}}) \\
\tau_m^{-1} &= \sigma([x_t, u_{t-1}] \mathbin\Vert W_{\tau_m}) \\
b_t &= \rho b_{t-1} + (1 - \rho) s_{t-1} \\
\theta_t &= 0.1 + 1.8 b_t \\
\Delta u &= (-u_{t-1} + x_t) / \tau_m \\
u_t &= u_{t-1} + \Delta u \\
s_t &= f_s(u_t, \theta) \\
u_t &= u_t (1 - s_t) + u_{\text{rest}} s_t
\end{align*}
where ${u}_{\text{rest}}$ denotes the neuron's resting potential and $\theta_t$ is the adaptive threshold modeled after adaptive spiking neurons ~\cite{yin2021accurate}, with $\tau_m$ and $\tau_{\text{adp}}$ representing the dynamically computed liquid time-constants.
\end{document}